\documentclass[letterpaper, 10pt, conference]{ieeeconf}  

\IEEEoverridecommandlockouts         
\usepackage{graphics} 
\usepackage{epsfig} 
\usepackage{mathptmx} 
\usepackage{times} 
\usepackage{amsmath} 
\usepackage{amssymb}  
\usepackage{multirow} 
\usepackage{xcolor}

\usepackage{bm}
\usepackage{booktabs}
\usepackage{multirow}
\usepackage{makecell}
\usepackage{nicefrac}
\usepackage{graphicx}
\usepackage{booktabs}
\usepackage{caption}
\usepackage{colortbl}
\usepackage{algorithm}
\usepackage{algorithmic}

\usepackage{graphicx}
\usepackage{subcaption}
\usepackage{graphicx}
\usepackage{comment}
\usepackage{thmtools,thm-restate}
\usepackage{wrapfig}
\usepackage{multirow}
\usepackage{xcolor}

\makeatletter
\let\NAT@parse\undefined
\makeatother
\usepackage[breaklinks,colorlinks,citecolor=green]{hyperref}

\definecolor{MyCyan}{RGB}{0,163,218}
\definecolor{MyDarkBlue}{RGB}{0,103,165}
\definecolor{MyDarkGreen}{RGB}{56,116,51}
\definecolor{MyMagenta}{RGB}{200,18,126}
\definecolor{irosblue}{RGB}{0, 0, 244}
\definecolor{irosred}{RGB}{244, 0, 0}
\definecolor{irosgreen}{RGB}{0, 244, 0}
\definecolor{lightgray}{gray}{0.8}

\newcommand\mypar[1]{\par\vspace{1.0mm}\noindent\textbf{#1}\;\;}
\title{\LARGE \bf
Joint Pedestrian Trajectory Prediction through Posterior Sampling
}

\author{Haotian Lin$^{1^*,2}$, Yixiao Wang$^{1^\dagger}$, Mingxiao Huo$^{3}$, Chensheng Peng$^{1}$, Zhiyuan Liu$^{2}$, Masayoshi Tomizuka$^{1}$
\thanks{$^{1}$ University of California, Berkeley, $^{2}$ Tsinghua University, $^{3}$ Carnegie Mellon University}
\thanks{
$^*$ Research performed while visiting University of California, Berkeley.}
\thanks{
$^\dagger$ Corresponding author: {\tt\small yixiao\_wang@berkeley.edu}}
}

\begin{document}

\maketitle
\thispagestyle{empty}
\pagestyle{empty}

\begin{abstract}
Joint pedestrian trajectory prediction has long grappled with the inherent unpredictability of human behaviors. Recent works employing conditional diffusion models in trajectory prediction have exhibited notable success. Nevertheless, the heavy dependence on accurate historical data results in their vulnerability to noise disturbances and data incompleteness. To improve the robustness and reliability, we introduce the Guided Full Trajectory Diffuser (GFTD), a novel diffusion-based framework that translates prediction as the inverse problem of spatial-temporal inpainting and models the full joint trajectory distribution which includes both history and the future. By learning from the full trajectory and leveraging flexible posterior sampling methods, GFTD can produce accurate predictions while improving the robustness that can generalize to scenarios with noise perturbation or incomplete historical data. Moreover, the pre-trained model enables controllable generation without an additional training budget. Through rigorous experimental evaluation, GFTD exhibits superior performance in joint trajectory prediction with different data quality and in controllable generation tasks. See more results at \url{https://sites.google.com/andrew.cmu.edu/posterior-sampling-prediction}.
\end{abstract}

\section{INTRODUCTION}
\label{sec:intro}
Pedestrian trajectory prediction is crucial for human-robot interaction systems such as autonomous driving, etc.  The goal is to predict future trajectories based on previous pedestrian movements and environmental contexts. By accurately predicting pedestrian trajectories, autonomous systems can plan their actions accordingly, ensuring safe and efficient navigation in dynamic environments.
However, the unpredictable and complicated nature of human behaviors makes this task challenging, especially with multiple agents involved where their interactions need to be considered.

To simplify the problem, early works \cite{yuan2021agentformer} \cite{wang2022stepwise} \cite{yue2022human} \cite{wong2022view} focused on marginal pedestrian trajectory prediction, which forecasts the trajectory for each pedestrian independently. Such approaches require a downstream planning module to perform safety checks for every combination of the individual predictions. Even so, combination rollouts could still produce unrealistic self-collisions and lead to failure in challenging scenarios. 
As a result, joint pedestrian trajectory prediction, which predicts consistent trajectories for all agents together, has gained attention in the community. \cite{weng2023joint} introduced a joint metrics term into supervision loss, transforming the marginal predictor into a joint one. However, joint pedestrian trajectory prediction remains challenging since existing approaches heavily rely on accurate and complete historical data to incorporate temporal and social dynamics. This dependence results in their vulnerability to interference from noisy disturbances and incomplete data, significantly threatening their effectiveness in real-world applications, such as sensors under adverse weather conditions.

\begin{figure*}
    \centering
    \includegraphics[width=0.95\textwidth]{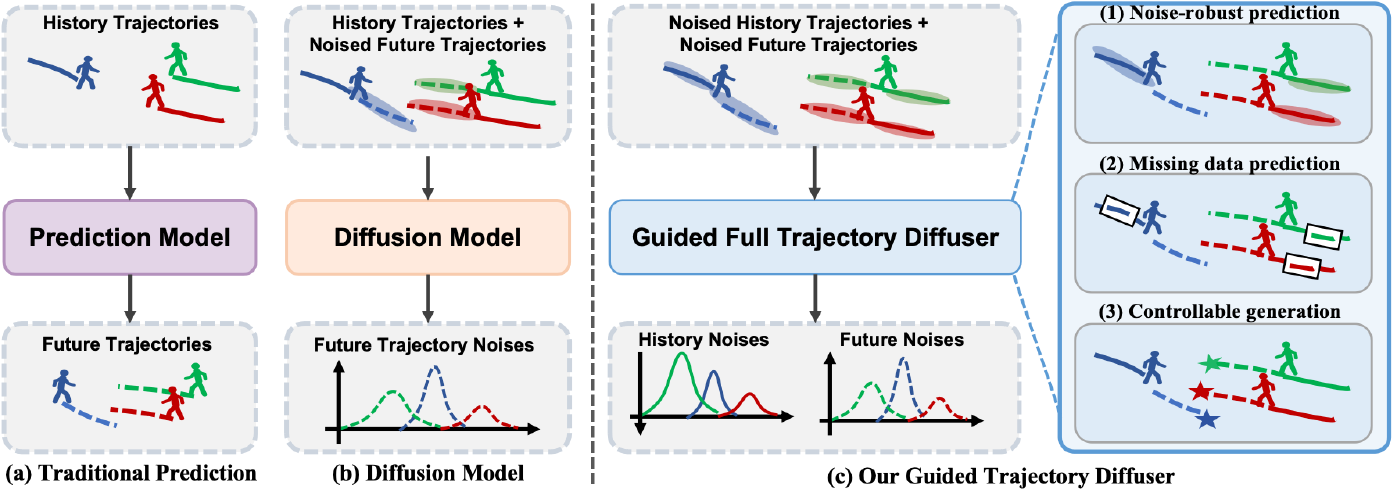}
    \caption{Illustration of existing trajectory prediction framework and our guided full trajectory diffuser framework. (a) Multi-agent trajectory prediction methods directly generate entire future trajectories through supervised learning. (b) Diffusion-based Multi-agent trajectory prediction methods generate future trajectories step-by-step during the denoising process. (c) Our method for multi-agent trajectory prediction incorporates history guidance within the diffusion framework. It predicts entire trajectories and supports additional capabilities in a single model, including Noise-Robust Trajectory Prediction, Incomplete Data Prediction, and Controllable Trajectory Generation.}
    \label{fig:comparison_with_previous_works}
\end{figure*}


To deal with the noises, previous research involved augmenting data with predefined noise and training models on noisy datasets~\cite{zamboni2022pedestrian}, and it can be further improved by adversarial training procedures \cite{cao2023robust}. In addition, historical trajectories from sensors could be incomplete. Previous studies ~\cite{cruz2019trajectory} \cite{wang2020reconstruction} \cite{qi2020imitative} \cite{xu2023uncovering} proposed to reconstruct incomplete data and predict future trajectories during the training phase, requiring a specifically designed model and well-established training strategy. Such approaches optimized per-problem functions, lacking adaptability to diverse tasks across various contexts.

Drawing inspiration from the diffusion model with its remarkable capability of capturing the complicated distribution, we propose a unified framework, named, \textbf{Guided Full Trajectory Diffuser} (\textbf{GFTD}) for the joint pedestrian trajectory prediction to better handle the disturbances and incompleteness.
  GFTD represents the entire trajectory distribution, both historical and future, with one diffusion model. We formulate trajectory prediction and controllable generation as inverse problems and solve them through posterior sampling techniques. Specifically, we sample in-distribution full trajectories based on historical trajectories and priors such as physical constraints and behavioral intentions. 
  Without the necessity for explicit training to handle noisy and/or incomplete inputs, 
  GFTD enables robust prediction and controllable generation—both achievable during the inference time. In a nutshell, our proposed framework streamlines the training process and offers adaptability to various scenarios at inference, providing a solution that can address all challenges without extra training requirements.
  

Our contributions can be summarized as follows:

\noindent  \textbf{(1)} We introduce a novel permutation-invariant framework for representing the joint distribution of full trajectories (both historical and future), converting trajectory prediction and controllable generation into a unified inverse problem.

\noindent \textbf{(2)} We utilize posterior sampling to solve the formulated problem. With our approach, there is no need for specific treatments during the training phase, as it can generalize to various types of data imperfections solely at inference time with one trained model.

\noindent \textbf{(3)} Extensive experiments demonstrate that our model not only performs comparably in joint trajectory prediction but also excels in controllable generation, particularly in scenarios with noise injection and incomplete historical data.

\section{Related Work}

\subsection{Pedestrian Trajectory Prediction}
Pedestrian trajectory prediction is crucial for many downstream tasks in autonomous driving, such as tracking \cite{peng2024pnas} and mapping \cite{peng2024q}. This task involves forecasting the future movement paths of pedestrians, given their historical movements and the environment. However, it is challenging to predict their motion because of the diverse and unpredictable nature of human behaviors. To address this issue, two mainstream paradigms have been developed. The supervised learning approach \cite{wong2022view} aims to minimize the differences between the ground truth trajectories and the predictions using L2 loss, etc.  On the other hand, the generative learning approach \cite{yuan2021agentformer} formulates the prediction as a task of generating conditional distributions of the future trajectory based on past trajectories. Among various generative models, the diffusion model \cite{gu2022stochastic} \cite{mao2023leapfrog} has shown exceptional performance in pedestrian trajectory prediction. Through a conditional reverse diffusion process, the diffusion model generates future trajectory distributions from a standard Gaussian distribution, which captures accurate and diverse predictions of future trajectories. 

\subsection{Joint Trajectory Distribution Modelling}
Recently, joint pedestrian trajectory prediction has gained significant attention. Unlike marginal trajectory prediction, which treats each pedestrian independently and can result in self-colliding trajectories between agents, joint trajectory prediction considers the interactions between future trajectories of agents, leading to more consistent and feasible predictions. \cite{weng2023joint} incorporates joint metrics as the training objective, transforming the marginal trajectory predictor into a joint trajectory predictor. In this paper, we propose Guided Full Trajectory Diffuser (GTFD) for joint pedestrian trajectory prediction, leveraging the strengths of diffusion models in generating accurate and diverse future trajectory distributions while considering the interactions between pedestrians.


\subsection{Posterior Sampling for Inverse Problems}
Posterior sampling is to infer the underlying distribution conditioned on the measurements, given the unconditional distribution. Previous works \cite{karras2022elucidating}  \cite{chung2022diffusion} \cite{boys2023tweedie} have demonstrated great performance in general inverse problems of computer vision, such as image inpainting and denoising.  Recent work involves physical constraints \cite{graikos2022diffusion}, the reward in reinforcement learning \cite{janner2022planning}, behavior preference in traffic simulation \cite{zhong2023guided} \cite{chen2024social} and bias in trajectory prediction \cite{jiang2023motiondiffuser}, extending a wide range of applications of posterior sampling. In this paper, we gain the insight that, in the task of trajectory prediction, historical trajectories can be regarded as observation or measurement. As a result, we can unleash the potential of posterior sampling in dealing with prediction under uncertainty. 

\section{Methods}
In this section, we undertake a comprehensive exposition of our novel multi-agent trajectory prediction framework, conceptualizing the prediction task through spatial-temporal inpainting. We first introduce preliminaries on diffusion models and problem definition. Subsequently, we explain how we formulate our Guided Full Trajectory Diffuser (GFTD).

\begin{figure}
    \centering
    \includegraphics[width=\linewidth]{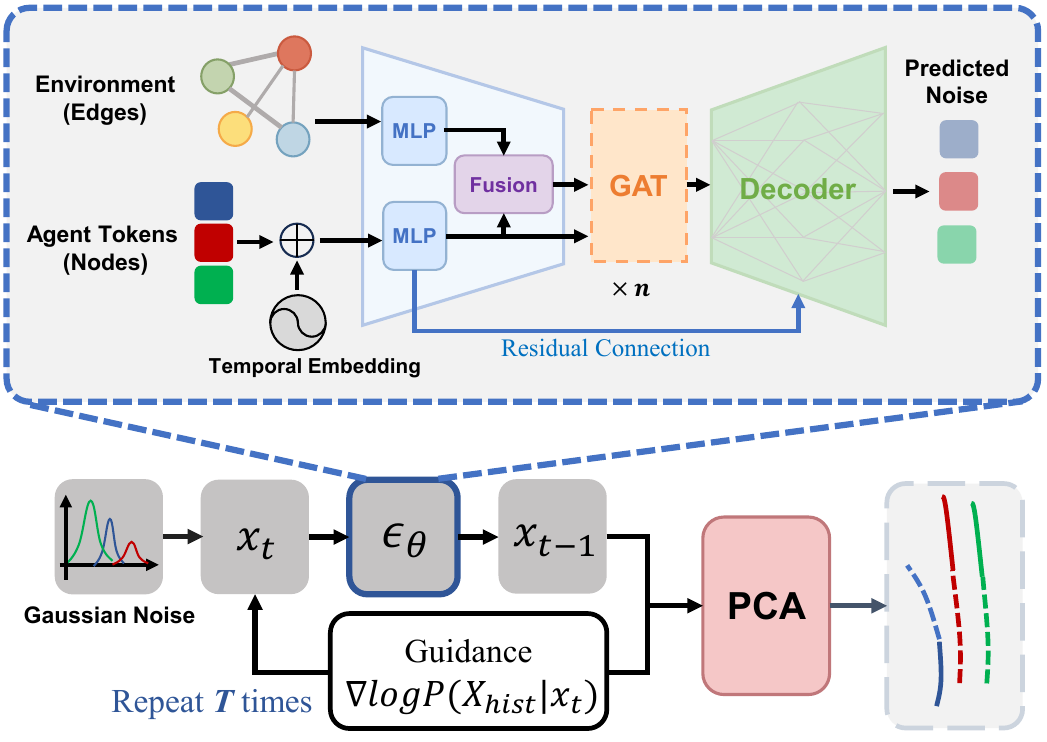}
    \caption{Architecture of our proposed GFTD framework. During inference, GFTD samples from Gaussian noise, and iteratively recover data with the denoise module $\epsilon_{\theta}$. It takes in noisy latent node features $x_t^i $ and edge features $e$. After MLP encoding, the edge is augmented by concatenating the encoded node features and they are both sent into the Processor which consists of stacked Graph Attention (GAT) layers. We then map the nodes to the same dimensions as $ x_t $ and add residual connection, resulting in the predicted intermediate noise $ \epsilon_t $. After each denoise step, we address conditions through posterior sampling. Finally, the denoised latent nodes $x_0$ are converted to trajectory space.}
    \label{fig:framework}
    
\end{figure}

\subsection{Preliminaries on Diffusion models}
Diffusion models are a class of generative models that operate on the principle of stochastic processes. They define a forward diffusion process that corrupts data by progressively introducing Gaussian noise. Conversely, in the reverse process, diffusion models reconstruct data from noise by iteratively reducing the Gaussian noise.

According to Itô Stochastic Differential Equations (SDE) \cite{song2020score}, the forward data noising process is defined as the following form:
\begin{equation}
    \label{eqn:forward sde}
    dx = f(x,t)dt+g(t)dw.
\end{equation}
where $x$ represents the data, $w$ is the standard Wiener process, $f$ is the drift coefficient, $g$ is the diffusion coefficient, $t\in [0,T]$ is the diffusion step.

The corresponding reverse SDE of Eq. (\ref{eqn:forward sde}) is defined as
\begin{equation}
    \label{eqn:reverse sde}
    dx = [f(x,t)-g^2(t)\nabla_{x}{\log{p_t}(x)}]dt+g(t)dw.
\end{equation}
where $p_t(x)$ is the probability density, and $\nabla_{x}{\log{p_t}(x)}$ is the score function, which can be learned by a neural network $s_{\theta}(x_t,t)$ with score matching. Here, $x_t$ denotes the intermediate noisy data $x$ at denoise time $t$. The training objective is defined as
\begin{equation}
    \label{eqn: score-matching}
    \mathbb{E}_t \{ \lambda(t)\mathbb{E}_{x_0}\mathbb{E}_{x_t | x_0}[\left\|s_{\theta}(x_t,t)-\nabla_{x_t}{\log{p}(x_t|x_0)}\right\|_2^2]\}.
\end{equation}
where $\lambda(t)$ is a weighting function. 

In this paper, we follow the implementation introduced by Denoising Diffusion Probabilistic Models (DDPM) \cite{ho2020denoising}. DDPM gradually add noise to the original data $x$ according to a variance schedule $\beta_1$, $\beta_2$,...,$\beta_T$:
\begin{equation}
    \label{eqn: diffuse}
    x_{t} = \sqrt{1-\beta_{t}}x_{t-1} + \sqrt{\beta}_{t} \epsilon, \quad t \in \{1,2,...,T\}
\end{equation}
where $\epsilon \sim \mathcal{N}(0,I)$, $x_0=x$, $x_T \sim \mathcal{N}(0,I)$. During the training, DDPM tends to learn the added noise, 
\begin{equation}
\label{eqn: DDPM training obj}
\mathbb{E}_{\epsilon,x_0,t}\left\|\epsilon-\epsilon_{\theta}(x_t,t)\right\|^2.
\end{equation}
With the learned $\epsilon_{\theta}(x_t,t)$, the reverse diffusion process gradually denoises standard Gaussian noise into the original data distribution by 
\begin{equation}
    \label{eqn: denoise}
    x_{t-1} = \frac{1}{\sqrt{\alpha_t}}(x_t-\frac{\beta_t}{\sqrt{1-\overline{\alpha}_t}}\epsilon_{\theta}(x_t,t))+\sqrt{\beta_t}\epsilon.
\end{equation}
where $x_T \sim \mathcal{N}(0,I)$, $t$ from $T$ to $1$.






\subsection{Problem Formulation}
The objective of the multi-agent trajectory prediction task is to use the observed history trajectories to jointly predict future trajectories of all the agents in the scene. We denote $P_{k}^i $ as the positions of agent $i$ at time step $k$, $i = 1, \dots, N_{a}$. Joint history trajectory is $c = \{c^i\}_{i=1}^{N_a}$, $c^i=\{ P_{- T_{his} + 1}^i, P_{- T_{his} + 2}^i, \dots\, P_{-1}^i\}$. Joint current position is $E = \{E^i\}_{i=1}^{N_a}$, $E^i=P_{0}^i$. JToint future trajectory is $y = \{y^i\}_{i=1}^{N_a}$, $y^i=\{ P_{1}^i, \dots, P_{T_{fut}}^{i} \}$. The task is to generate the future trajectory distribution $p(y|E, c)$. However, the historical trajectory may be incomplete and injected with noise caused by unexpected conditions such as sensor failure and poor weather. We use an elementwise mask to represent incomplete trajectory data and model the injected noise as Gaussian noise.

\subsection{Represent Full Trajectory Distribution with Diffusion}
Previous diffusion-based work \cite{gu2022stochastic} \cite{ jiang2023motiondiffuser} \cite{wang2024optimizing} formulated prediction task as the diffusion process that conditioned on history trajectories $c$ and current position $E$ to generate future trajectory distribution $p(y|E, c)$. Our key insight here is that we can learn to generate full-length trajectories containing both the past and the future and regard the prediction task as an inverse problem of inpainting, which aims to infer and complete the incomplete trajectory (e.g., the future) based on observations (e.g., the past). We denote the generated history trajectory as $\hat{c}_i$ and the corresponding full-length trajectory as $x^i = \hat{c}^i \cup y^i  = \{ P_{- T_{his}}^i, \dots, P_{-1}^{i}, P_{1}^{i}, \dots, P_{T_{fut}}^{i} \}$. We train the DDPM  by Eq. (\ref{eqn: DDPM training obj}) where $x = \{x^i\}_{i=1}^{N_a}$, and we can generate full trajectory distribution $p(x|E)$ by learned $\epsilon_\theta(x_t,t,E)$ through Eq. (\ref{eqn: denoise}).

\subsection{Robust Prediction as Posterior Sampling}
Prediction task is to obtain $p(y|E,c)$ which is equivalent to $p(y,c|E,c)=p(x|E,c)$. Given learned full trajectory $p(x|E)$ through diffusion model and Bayes' rule, we have $p(x|E, c)=p(c|x, E)p(x | E)/p(c)$. Thus, the score of the posterior distribution can be calculated as:
\begin{equation}
     \nabla_{x}{\log{p_t}(x|E,c)}= \nabla_{x}{\log{p_t}(x|E)}+\nabla_{x}{\log{p_t}(c|x, E)}.
\end{equation}
According to diffusion posterior sampling (DPS) \cite{chung2022diffusion}, if the condition or measurement has the form of a general noisy inverse problem $c=\phi(x_0) + n$, where $\phi$ is an arbitrary operator, and $n \sim \mathcal{N}(0, \sigma ^2I)$ is the Gaussian noise, then $\nabla_{x}{\log{p_t}(c|x, E)}$ can be approximated as
\begin{equation}
     \label{eqn: guidance term}
     \nabla_{x}{\log{p_t}(c|x, E)}=-\lambda \nabla_{x_t}\left\|c-\phi(\hat{x_0}(x_t))\right\|_2^2 = -\lambda \nabla_{x_t}\mathcal{L},
\end{equation}
where $\hat{x_0}(x_t)$ is the posterior mean of $p(x_0|x_t)$:
\begin{equation}
     \hat{x_0}(x_t)=\frac{1}{\sqrt{\overline{\alpha}_t}}(x_t+(1-\overline{\alpha}_t)\epsilon_{\theta}(x_t,t,E)).
\end{equation}

Once the condition operator $ \phi $ is known (i.e., $\phi_{his} (x)$ denotes historical portion of $x$), we can inject such condition by iteratively adding the guidance term Eq. (\ref{eqn: guidance term}) to the intermediate noisy data during the reverse diffusion process so that we drag the sample to the direction that minimizes the guidance loss $ \mathcal{L} $ that reflects our preferences, which would be having the generated history trajectories close to the ground truth observation. Specific designs of $ \mathcal{L} $ will be elaborated in section \uppercase\expandafter{\romannumeral4}. With the guidance gradient restricted to a certain range, the generated noisy samples will be lying in the data manifolds, thus avoiding generating out-of-distribution samples. Even though this does not guarantee the exact recovery of history trajectories and may result in some degrees of ill-conditioned prediction, a properly guided sample still achieves high prediction accuracy. We refer to this characteristic as "soft-conditioning" and will show such a design greatly boosts robustness and adaptation abilities in the following sections. 

For conventional prediction tasks, we expect data to be clean and complete and do not need to trade condition correctness off for perturbation tolerance. To further enhance performance on these tasks, we can strictly enforce history conditions following RePaint\cite{lugmayr2022repaint}. This plug-and-play modification can be manually enabled. It introduces an additional step that first compromises observed history trajectories to the noise level $t$ through the forward diffusion process~Eq. (\ref{eqn: diffuse}), then concatenates them with the corresponding future parts of intermediate samples.

An implementation of the inference process in our proposed framework is summarized in Algorithm~\ref{alg: GFTD}. 

\begin{algorithm}
	\renewcommand{\algorithmicrequire}{\textbf{Input:}}
	\renewcommand{\algorithmicensure}{\textbf{Output:}}
	\caption{Guided Full Trajectory Diffuser (GFTD)}
	\label{alg: GFTD}
	\begin{algorithmic}[1]
        \REQUIRE $\mathcal{L}(\cdot, \cdot)$, $\phi(\cdot)$, $c$, $\{\Bar{\alpha}_t, \beta_t,\lambda_t\}_{t=0}^{T-1}$,
        
        \STATE $x_T \sim \mathcal{N}(0,I)$
        \FOR {$t=T-1,...,1$}
            \STATE $\epsilon \sim \mathcal{N}(0,I)$
            \STATE $x_{t-1} = \frac{1}{\sqrt{\alpha_t}}(x_t-\frac{\beta_t}{\sqrt{1-\overline{\alpha}_t}}\epsilon_{\theta}(x_t,t, E))+\sqrt{\beta_t}\epsilon$ 
            \STATE $\hat{x}_0 = \frac{1}{\sqrt{\Bar{\alpha}_t}}(x_t-\sqrt{1-\Bar{\alpha}_t} \epsilon_\theta(x_t,t,E))$
            \STATE $g=-\nabla_{x_{t}}\mathcal{L}(\hat{x}_0, c)$
            \STATE $x_{t-1} = x_{t-1} + \lambda_t g
            $ 
            \IF{RePaint \AND $t > 0$}
            \STATE $\epsilon' \sim \mathcal{N}(0,I)$
            \STATE $c_{t-1} = \sqrt{1-\beta_{t-1}}c_0 + \sqrt{\beta}_{t-1}
            \epsilon$
            \STATE $\phi_{his}(x_{t-1}) = c_{t-1}$
            \ENDIF
        \ENDFOR
        \STATE $ y=\phi_{fut}(x_0) $
		\ENSURE $y$
	\end{algorithmic}  
\end{algorithm}

\subsection{Trajectory Latent Representation}
Inspired by the advances in image synthesis bought by latent diffusion models \cite{rombach2022high} and the successful application of low-rank latent representation in trajectory prediction \cite{bae2023eigentrajectory} \cite{jiang2023motiondiffuser}, we further consider transforming the spatial-temporal features of trajectory $X^i \in \mathbb{R}^{T * d}$ into latent features $x^i\in \mathbb{R}^{k}$, $k \ll T*d$ through Principal Component Analysis (PCA). This notation stands for the linear combination of the first principal components. This simple but efficient transformation serves to mitigate the effects of noisy data, leading to more consistent and smoother predicted trajectories. Meanwhile, the latent representation contains its geometry and temporal characteristics, saving the need for additional temporal encoder blocks.

\section{Framework Architecture}
As we re-formulate the prediction task as trajectory inpainting, our proposed framework requires a pre-trained DDPM that models the joint distribution of full-length trajectories that is not conditioned on history motion $c$. Through iteratively adding posterior guidance $-\lambda \nabla_{x_t}\mathcal{L}$ during inference time and extracting the future parts $y$ from generated trajectories $x$, we achieve the goal to sample future trajectories conditioned on given histories. We argue that our framework is model-agnostic that various reasonable network designs of the denoising module are feasible for our methodology. This section presents a specific implementation of a lightweight denoising module. We will discuss our data representation method, module architecture, and guidance design in the following.

\mypar{Data Representation.} To achieve a rotation-invariant representation of agent motion, for each agent in the scene, we normalized their trajectory according to their current positions and headings. Parallelly, we extract the spatial relativity of agents under current frame by establishing a graph representation, where node $x^i, i=1,\dots, N_a$ stands for agent motion and edge $E^{ij}, i,j=1,\dots, N_a$ is characterized by a 6D vector describing their relative position at the current timestep: relative distance $d^{ij}$, direction vector $r^{ij}$ under the reference frame of agent $i$, and relative heading vector $h^{ij} = \{\theta^{ij}, cos\theta^{ij}, sin\theta^{ij}\}$. 

\mypar{Denoise Module.} Following the standard setting in DDPM, we built a denoise module $\epsilon_{\theta}(x_t, t, E)$ that predicts the intermediate noise level and denoise the intermediate sample $x_t$ by Eq. (\ref{eqn: denoise}). For each scenario sample, $x_t = [PCA(X_t^1), \dots, PCA(X_t^{N_a})]$ is the latent node features of agent trajectories at diffusion step $t$. And $E$ stands for the edge information mentioned above, it serves as the only condition of the diffusion model and is shared across all diffusion steps.

The core to $\epsilon_{\theta}$ is a Graph Neural Network (GNN). As depicted in Fig.~\ref{fig:framework}, each latent node feature $x_t^i $ is firstly concatenated with diffusion time embeddings and then mapped to higher dimensions by a multi-layer perception (MLP). Since node features do not contain global position information, we need to further fuse the nodes with edge features so that we can model the complex interaction. After MLP encoding, the edge is augmented by concatenating with the encoded node
\begin{equation}
    E_t^{ij} = concat\{ E^{ij}, x_t^{i} \},
\end{equation}
and they are both sent into the Processor which consists of stacked Graph Attention (GAT) \cite{velivckovic2017graph} layers. Within each GAT layer, nodes are augmented by shared linear transformation parameterized by weight $\boldsymbol{W}$. Then, we calculate the attention score by
\begin{equation}
    \alpha_t^{ij} = softmax_{j}(e_t^{ij}) 
    = \frac{exp(e_t^{ij})}{\sum_{k \in N_a}exp(e_t^{ik})},
\end{equation}
and
\begin{equation}
    e_t^{ij} = LekyReLU(a(\boldsymbol{W}n^i || \boldsymbol{W}n^j)),
\end{equation}
where $a(\cdot)$ linear transformation that maps a vector to a real number. After acquiring interaction information, we map the nodes to the same dimensions as $ x_t $, resulting in the predicted intermediate noise $ \epsilon_t $.

\mypar{Guided Generation.} For trajectory prediction under the description of the noisy inverse problem, the closed-form dependency between the measurement and sample can be formulated as
\begin{equation}
c = \phi(X) + n.
\end{equation}
For the prediction task, $c$ is the observed history trajectories, $\phi$ is the operator that extracts history parts of trajectories from full-length trajectories $X$, and $n$ is the Gaussian noise with adjustable variance based on how clear our history data is. For the original prediction task, we consider the history trajectories to be reliable and the noise level to be low. Similar to the derivation in section $C$, we set our guidance loss as
\begin{equation}
    \label{reconstruction loss}
    \mathcal{L}_{rec} = \left\|c-\phi(\hat{X}_0(x_t))\right\|_2,
\end{equation}
which is the reconstruction loss that measures how precisely our generated trajectories fit the conditions. Inspired by \cite{jiang2023motiondiffuser}, we can ensure a more realistic generation by adding repeller guidance that prevents the agents from colliding with each other:
\begin{equation}
    \mathcal{L}_{rep} = \frac{1}{N_a}\sum_{i, j, k} \mathrm{max}\{(1 - \frac{1}{r}d^{ij}_k),0\},
\end{equation}
where $r$ is the repeller threshold and $ d^{ij}_k \in \mathbb{R}^{N_a \times N_a \times (T_{hist} + T_{fut})} $ is the distance between agent $i$ and $j$ at timestep $t$. Practically, the repeller loss serves as valuable prior knowledge.


\section{Experiments}
 In this section, we exhibit the capacity and flexibility of our framework by adapting our model to four distinct tasks without retraining: basic trajectory prediction, controllable generation, prediction with noisy history, and prediction with incomplete history. We present two versions of our proposed framework: the foundational \textbf{GFTD} and its variant, \textbf{GFTD-RePaint}, which is specialized for the basic trajectory prediction task.

\begin{figure}
    \centering \includegraphics[width=\linewidth]{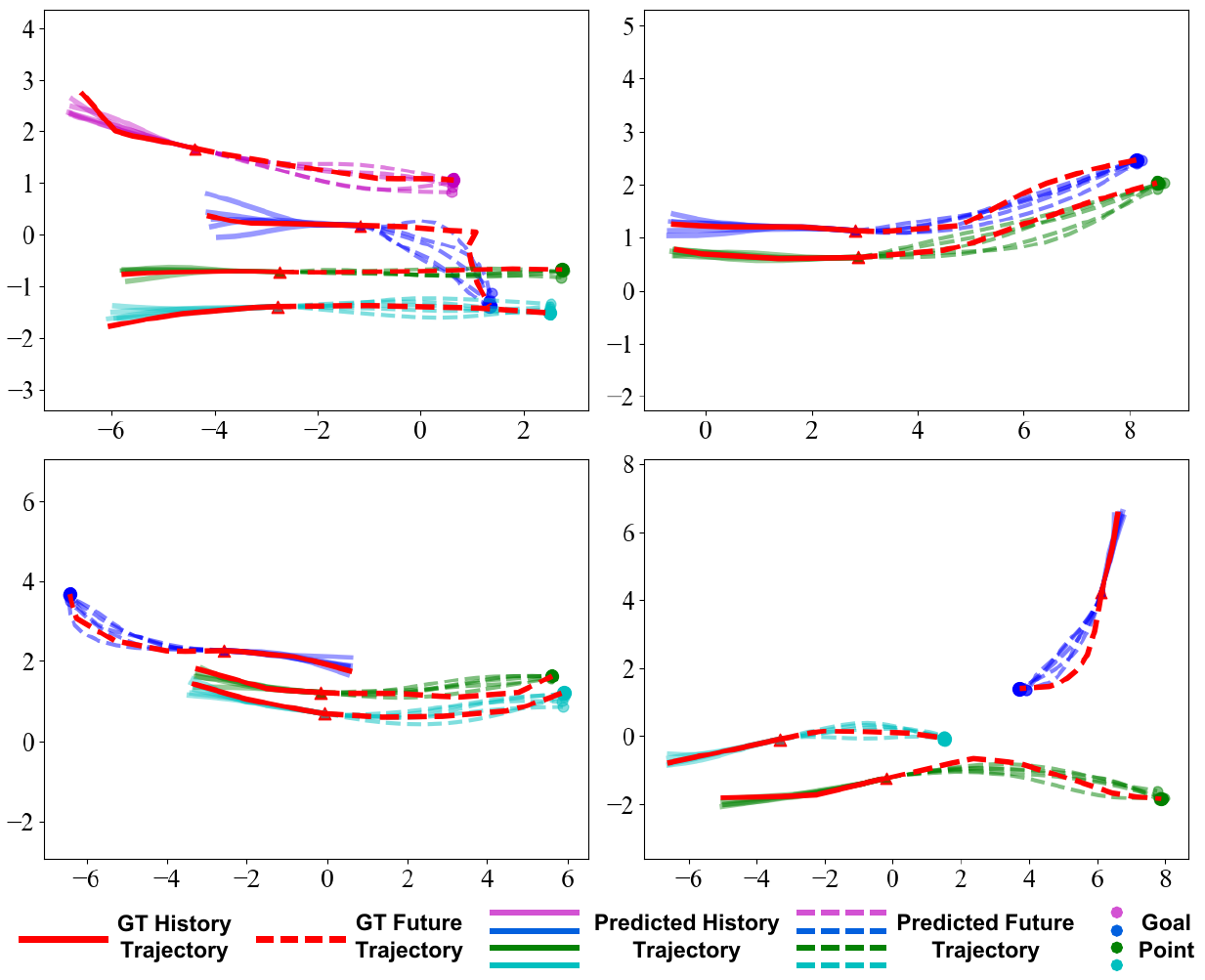}
    \caption{Qualitative visualization of controllable generation. Red lines represent the ground truth trajectories. With the given goal point and history, our model can generate goal-oriented trajectories with considerable realism.}
     \label{fig: controllablegeneration}
\end{figure}

\begin{table*}[]
\caption{Trajectory Prediction Performance on ETH/UCY Dataset. 
The three best scores are marked by \textcolor{irosred}{red}, \textcolor{irosblue}{blue}, and \textcolor{irosgreen}{green}, respectively. $\dagger$ denotes Joint AgentFormer without the diverse sampler (DLow)}
\label{table: prediction}
\centering
\begin{tabular}{lllllll}
 \toprule
\multirow{2}{*}{\textbf{Method}}  & \multicolumn{6}{c}{\textbf{minJADE~/~JFDE~(m), K=20} $\downarrow$} \\  
\cmidrule{2-7}
& {ETH(1.4)}       & {HOTEL(2.7)}             & {UNIV(25.7)}    & {ZARA1(3.3)}    & {ZARA2(5.9)}     & ETH/UCY Avg.  \\ \midrule
S-GAN \cite{gupta2018social}                & {0.919 / 1.742} & {0.480 / 0.950}          & {0.744 / 1.573} & {0.438 / 1.001} & {0.362 / 0.794}  & 0.589 / 1.212 \\ 
PECNet \cite{daniel2021pecnet}            & {0.618 / 1.097} & {0.291 / 0.587}          & {0.666 / 1.417} & {0.408 / 0.896} & {0.372 / 0.840}  & 0.471 / 0.967 \\ 
MemoNet\cite{xu2022remember}              & {\textcolor{irosred}{0.499} / \textcolor{irosblue}{0.859}} & {0.222 / 0.416}          & {0.686 / 1.466} & {0.349 / 0.723} & {0.385 / 0.864}  & 0.428 / 0.866 \\
Joint View Vertically \cite{weng2023joint}   & {0.652 / \textcolor{irosred}{0.839}} & {\textcolor{irosblue}{0.186} / \textcolor{irosblue}{0.309}}          & {\textcolor{irosred}{0.523} / \textcolor{irosred}{1.091}} & {\textcolor{irosgreen}{0.331} / \textcolor{irosblue}{0.634}} & {\textcolor{irosred}{0.267} / \textcolor{irosred}{0.547}}  & \textcolor{irosgreen}{0.392} / \textcolor{irosred}{0.684} \\
 Joint AgentFormer$^\dagger$ \cite{weng2023joint}   & {0.543 / 0.883} & {0.211 / 0.377}   & {\textcolor{irosblue}{0.596} / \textcolor{irosblue}{1.247}}  &  {\textcolor{irosred}{0.309} / \textcolor{irosred}{0.612}} & {\textcolor{irosblue}{0.282} / \textcolor{irosblue}{0.584}} &    {\textcolor{irosblue}{0.388} / \textcolor{irosblue}{0.741}}      \\
Ours (GFTD)  & {\textcolor{irosblue}{0.505} / \textcolor{irosgreen}{0.873}} & {\textcolor{irosred}{0.174} / \textcolor{irosred}{0.297}} & {0.649 / 1.305}          & {0.340 / 0.667} & {0.308 / 0.620}  & {0.395 / 0.752}              \\
Ours (GFTD + RePaint)     & {\textcolor{irosgreen}{0.514} / 0.906} & {\textcolor{irosgreen}{0.191} / \textcolor{irosgreen}{0.329}}          & {\textcolor{irosgreen}{0.607} / \textcolor{irosgreen}{1.248}} & {\textcolor{irosblue}{0.327} / \textcolor{irosgreen}{0.646}} & {\textcolor{irosgreen}{0.288} / \textcolor{irosgreen}{0.585}} & {\textcolor{irosred}{0.385} / \textcolor{irosgreen}{0.743}} \\ 
\bottomrule
\end{tabular}
\end{table*}

\begin{table*}[]
\caption{Controllable Generation Performance}
\centering
\begin{tabular}{cllllll}
 \toprule

\multirow{2}{*}{\textbf{Method}}  & \multicolumn{6}{c}{\textbf{minJADE~/~JFDE~(m), K=20} $\downarrow$} \\
\cmidrule{2-7} 
& {ETH(1.4)}       & {HOTEL(2.7)}             & {UNIV(25.7)}    & {ZARA1(3.3)}    & {ZARA2(5.9)}     & ETH/UCY Avg.  \\ 
\midrule
Baseline  & {0.505 / 0.873} & {0.174 / 0.297} & {0.649 / 1.305}              & {0.340 / 0.667} & {0.308 / 0.620} &     {0.395 / 0.752}                        \\
Goal Point Guidance   & {0.224 / 0.064 } & {0.069 / 0.033 }      & {0.280 / 0.394  } & {0.103 / 0.035 } & {0.094 / 0.032 } &             {0.154 / 0.112}     
\\  \bottomrule
\label{table: Control}
\end{tabular}
\end{table*}

\subsection{Experimental Setups}
\mypar{Datasets.} We carried out all of our experiments on the ETH/UCY \cite{lerner2007crowds} \cite{pellegrini2009you} dataset, a popular public pedestrian trajectories forecasting benchmark. The dataset contains pedestrian trajectories in bird's eye view (BEV) from five distinct scenarios (ETH, Hotel, Univ, Zara1, Zara2). We follow the leave-one-out training/evaluation setup that was used in the original S-GAN \cite{gupta2018social}. Specifically, we partitioned each dataset scene into sliding windows of length 20 steps (8 seconds) at stride 1, with 8 history observation steps (3.2 seconds) followed by 12 prediction steps (4.8 seconds). We then proceed to utilize all sequences containing at least one pedestrian.

\mypar{Metrics and Baselines.}
For scene-level joint prediction, rather than using marginal metrics Average Displacement Error (ADE) and Final Displacement Error (FDE), we follow \cite{weng2023joint} to use joint metrics JADE/JFDE:
\begin{equation}
    \label{JADE}
    joint ADE(Y, Y^*) = \frac{1}{TN}\mathop{\mathrm{min}}\limits_{k=1}^{K}\sum\limits_{n=1}^{N}\sum\limits_{t=1}^{T} \left\| s_{t,n} - {s}_{t,n}^{*}\right\| ,
\end{equation}
\begin{equation}
    \label{JFDE}
    joint FDE(Y, Y^*) = \frac{1}{TN}\mathop{\mathrm{min}}\limits_{k=1}^{K}\sum\limits_{n=1}^{N} \left\| s_{T,n} - {s}_{T,n}^{*}\right\| .
\end{equation}
The joint metrics emphasize the significance of modeling the interactions between agents by introducing a constraint for top-$K$ evaluations: predictions must originate from the same sample. This approach prevents the overestimation of model performance and penalizes models that fail to accurately simulate the realistic behaviors and motions of all agents present in the scene. In the general prediction task, we compared our model to the state-of-the-art models Joint AgentFormer and Joint View Vertically proposed in \cite{weng2023joint}, as well as other famous baselines in pedestrian prediction: S-GAN \cite{gupta2018social}, PECNet\cite{mangalam2020not}, and MemoNet \cite{xu2022remember}. We compute the best-of-20 JADE/JFDE considering the stochasticity of generative prediction models.

\mypar{Implementation Details.} The implementation of the denoising module used in experiments contains a three-layer GAT with a hidden size of 128. Additionally, each feed-forward network is realized as a two-layer MLP equipped with post-layer normalization and activated using the Mish activation function \cite{misra2019mish}. In addition, we observed that training our diffusion model on latent space by Eq. (\ref{eqn: DDPM training obj}) can be unstable and somehow inefficient. Thus, following \cite{hang2023efficient}, we modified the DDPM training loss with the Min-SNR-$\gamma$ weighting strategy $ \lambda(t) = min\{ \gamma, SRN(t)\} / SRN(t) $ to avoid having the model putting too much attention on the final steps of the denoising processing where the noise level is low. We set the training batch size to 32 and we use the Adam optimizer with an initial learning rate of 0.001. All the training is conducted on one GTX-2080Ti GPU.

\subsection{Trajectory Prediction and Controllable Generation}
In the context of basic pedestrian trajectory prediction tasks, we present two models within our Guided Trajectory Diffusion (GFTD) framework: the GFTD model and its augmented version, GFTD with RePaint. To assess their efficacy, we conduct a quantitative evaluation, benchmarking these models against the aforementioned baselines with the ETH/UCY dataset. All models are fairly evaluated by joint metrics to ascertain their optimal scene-level performance across 20 samplings. We directly used the reported baseline performance under joint metrics from \cite{weng2023joint}. We present the results in TABLE \ref{table: prediction}. In specific for Jonit AgentFormer, we use the pre-trained checkpoint for joint prediction provided by their official GitHub repository (https://github.com/ericaweng/joint-metrics-matter). For a fair comparison, we did not apply the Dlow model for performance enhancement. Our model exhibited relative competitive performance, especially on the HOTEL dataset, where our model outperformed all the baselines.

Controllable trajectory generation is another plausible application of our framework. In addition to the fundamental reconstruction loss Eq. (\ref{reconstruction loss}), we can incorporate any desirable objectives into the guidance term to control the sampling process. For demonstration, we experimented with goal point generation, i.e., given the ground truth history trajectories and desired goal points $g$, we asked our model to generate future trajectories that reach the goal points. The only modification to adapt our pre-trained model to this particular task is to add an attraction term that measures the L2-norm between the endpoints of generated trajectories and goal points.

We evaluated generation quality through JADE, which represents the realism of the generated scene, and JFDE, which reflects goal-reaching accuracy. As shown in Fig~\ref{fig: controllablegeneration}, all samples tend to accurately recover the demanded goal point while the generated trajectories are smooth and realistic. The variance of prediction samples tends to be large under cases where the pedestrian performs a sudden turn. However, in most cases, the model successfully covers the modality that most resembles the ground truth. We also present quantitive results in TABLE \ref{table: Control}, which showcases the effectiveness of goal point guidance.


\begin{table*}[]
\caption{Prediction with Noisy Data}
\centering
\begin{tabular}{cccccccc}
\toprule
\multirow{2}{*}{\textbf{Noise Std}}&\multirow{2}{*}{\textbf{Model}} & \multicolumn{6}{c}{\textbf{minJADE~/~JFDE~(m), K=20} $\downarrow$} \\ 
\cmidrule{3-8}
& & {ETH(1.4)}       & {HOTEL(2.7)}             & {UNIV(25.7)}    & {ZARA1(3.3)}    & {ZARA2(5.9)}     &{ETH/UCY Avg.}  \\ 
\midrule

\multirow{2}{*}{0.00}&Joint AgentFormer \cite{weng2023joint} & {0.543 / 0.883} & {0.211 / 0.377}   & {\textbf{0.596} / \textbf{1.247}}  &  {\textbf{0.309} / \textbf{0.612}} & {\textbf{0.282} / \textbf{0.584}} &    {\textbf{0.388} / \textbf{0.741}}       \\

&Ours & {\textbf{0.505} / \textbf{0.873}} & {\textbf{0.174} / \textbf{0.297}}        & {0.649 / 1.305}   &  {0.340 / 0.667} &   {0.308 / 0.620}   &  0.395 / 0.752        \\   
\midrule

\multirow{2}{*}{0.05}&Joint AgentFormer \cite{weng2023joint} & {0.588 / \textbf{0.992} } & {0.253 / 0.433 }      & {\textbf{0.628} / \textbf{1.294} } & {\textbf{0.362} / \textbf{0.708} } & {0.361 / 0.714 } &    {\textbf{0.438} / 0.834 }       \\

&Ours & {\textbf{0.567} / 1.009} & {\textbf{0.198} / \textbf{0.334}} & {0.703 / 1.388 }              & {0.388 / 0.735} & {\textbf{0.341} / \textbf{0.675}} &   {0.439 / \textbf{0.828} }           \\   
\midrule

\multirow{2}{*}{0.15}&Joint AgentFormer \cite{weng2023joint} & {0.765 / 1.287 } & {0.458 / 0.762  }      & {\textbf{0.834} / 1.642  } & {0.649 / 1.254 } & {0.736 / 1.312 } &  {0.688 / 1.254 }         \\

&Ours & {\textbf{0.636} / \textbf{1.076}} & {\textbf{0.283} / \textbf{0.463}} & {0.872 / \textbf{1.661}}             & {\textbf{0.633} / \textbf{1.137}} & {\textbf{0.469} / \textbf{0.869}} &    {\textbf{0.579} / \textbf{1.041}}           \\  
\bottomrule
\label{table: noisy data}
\end{tabular}
\end{table*}

\begin{table*}[]
\caption{Prediction with Incomplete History Data}
\centering
\begin{tabular}{cccccccc}
\toprule
\multirow{2}{*}{\textbf{Missing ratio}}&\multirow{2}{*}{\textbf{Model}} & \multicolumn{6}{c}{\textbf{minJADE~/~JFDE~(m), K=20} $\downarrow$} \\ 
\cmidrule{3-8}
& & {ETH(1.4)}  & {HOTEL(2.7)}             & {UNIV(25.7)}    & {ZARA1(3.3)}    & {ZARA2(5.9)}     & {ETH/UCY Avg.}  \\ 
\midrule

\multirow{2}{*}{25\% } &Joint AgentFormer \cite{weng2023joint}  & {0.558 / 0.901} & {0.214 / 0.377} & {\textbf{0.624} / \textbf{1.286}} & {\textbf{0.336} / \textbf{0.653}} & {\textbf{0.303} / \textbf{0.614}} &       {0.407 / 0.766}          \\  

& Ours  & {\textbf{0.524} / \textbf{0.910}} & {\textbf{0.174} / \textbf{0.294}} & {0.662 / 1.321} & {0.341 / 0.670} & {0.314 / 0.627}   &   {\textbf{0.403} / \textbf{0.764}}    \\  
\midrule

\multirow{2}{*}{50\% } &Joint AgentFormer \cite{weng2023joint} & {0.619 / 1.022} & {0.219 / 0.378} & {\textbf{0.662} / \textbf{1.343}}              & {0.374 / 0.721} & {0.327 / 0.649} &       {0.440 / 0.823}          \\

& Ours  & {\textbf{0.511} / \textbf{0.897}} & {\textbf{0.176} / \textbf{0.302}} & {0.676 / 1.348}       & {\textbf{0.347} / \textbf{0.677}} & {\textbf{0.321} / \textbf{0.642}}  &   {\textbf{0.406} / \textbf{0.773}}       \\ 
\midrule

\multirow{2}{*}{75\% } &Joint AgentFormer \cite{weng2023joint} & {0.647 / 1.034} & {0.274 / 0.444} & {0.752 / 1.488} & {0.440 / 0.830} & {0.381 / 0.733} &       {0.499 / 0.906}          \\

& Ours   & {\textbf{0.519} / \textbf{0.885}} & {\textbf{0.193} / \textbf{0.333}} & {\textbf{0.735} / \textbf{1.441}} & {\textbf{0.357} / \textbf{0.696}} & {\textbf{0.346} / \textbf{0.683}} &        {\textbf{0.430} / \textbf{0.808}}         \\ 
\bottomrule
\label{table: missing data}
\end{tabular}
\end{table*}

\subsection{Prediction with Noisy History}
To assess the robustness of our proposed method against noisy input, we perturbed the observed history trajectories by introducing random Gaussian noise $n \sim \mathcal{N}(0,\sigma^2 I)$ of varying standard deviation $\sigma$. Specifically, we introduced two levels of perturbation: a slight level, characterized by a standard Gaussian noise with a 0.05-meter $\sigma$, and a heavy level, with a standard Gaussian noise with a 0.15-meter $\sigma$. Subsequently, we compared the performance of our model with the Joint AgentFormer baseline under identical experimental conditions.

\begin{figure}[t]
    \centering
    \includegraphics[width=\linewidth]{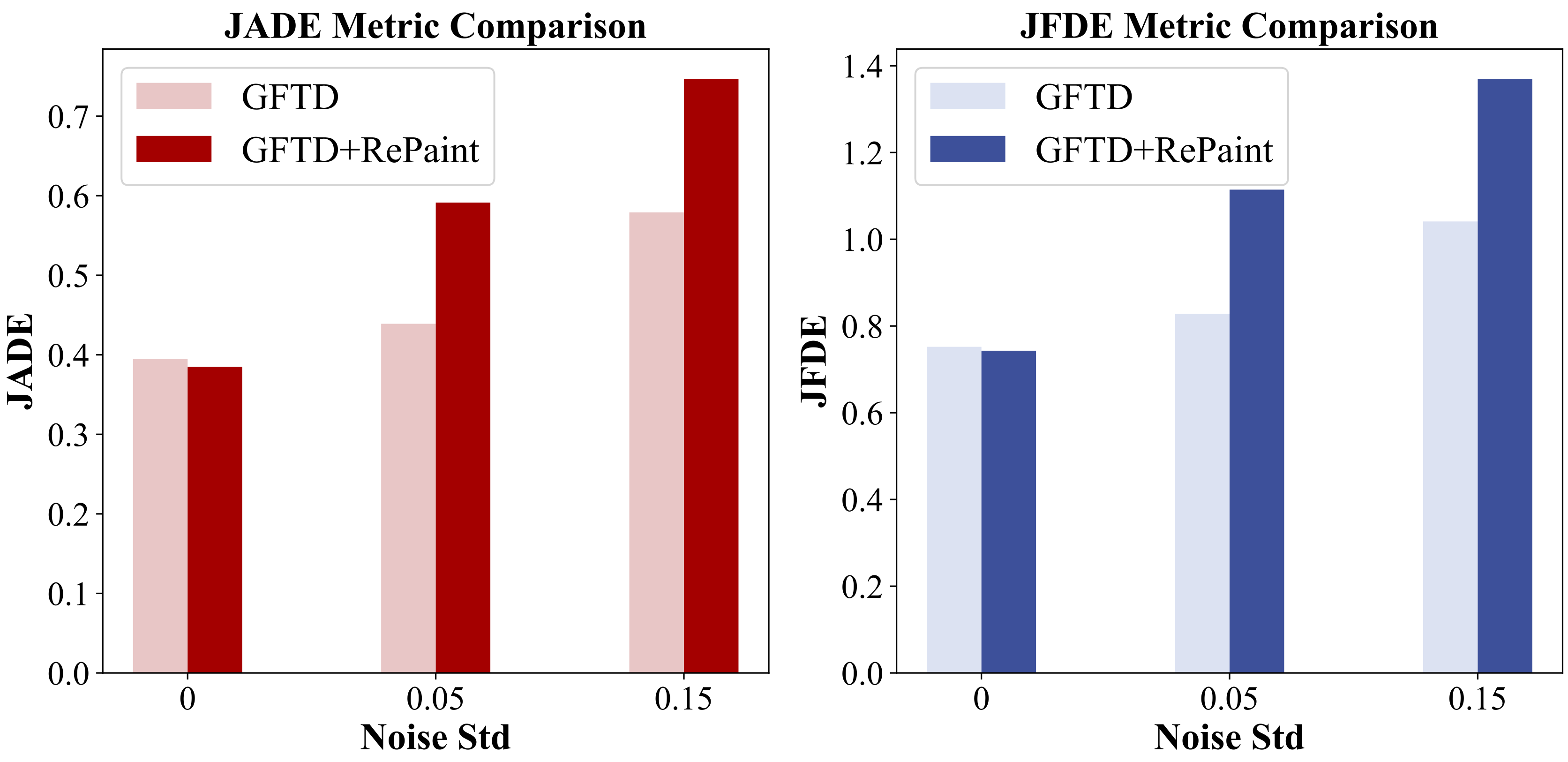}
    \caption{JADE/JFDE performance comparison between GFDT and GFDT+RePaint with noisy data input.}
    \label{fig:fig4}
\end{figure}

As shown in TABLE~\ref{table: noisy data}, we observed that our model performs better at heavy noise levels, providing trustworthy prediction even if the data is highly corrupted. The average JADE/JFDE performance of our method on 5 benchmarks is close to the baseline but only degenerated by $46.58\%/38.38\%$ at the heavy level. By contrast, the performance of the baseline method had degenerated by $57.08\% / 50.36\%$ at heavy noise level. To reasonably state that the guided posterior sampling accounted for the robustness against noisy data, we further compared it with GFTD-RePaint. GFTD-RePaint includes gradually replacing the history part of the intermediate noisy sample by ground truth history observation, thus ensuring the generated samples were strictly conditioned on the noisy history data. The experimental results in Fig.~\ref{fig:fig4} show that despite having a near performance to GFTD model under clean data conditions, the GFTD-RePaint performs much worse on noisy data, especially under heavy-level noise. It shows the superiority of our proposed framework in noisy conditions rooted in the design of "soft conditioning". 

\subsection{Prediction with Incomplete History}

Due to obstacles or sensor failures, historical data may occasionally contain unexpected missing frames. To evaluate model robustness against incomplete data, we randomly select and mask $25\%$, $50\%$, and $75\%$ of the historical trajectory frames, while ensuring that the current frame is always retained. With a known random mask, we only consider the available historical frames when calculating the guidance loss. We employ the same Joint AgentFormer model as our baseline. To ensure a fair comparison, we generate an attention mask for each AgentFormer sub-module based on the known data mask and replace the masked frames with zero values. For both models, we directly re-used the same pre-trained versions as in the other two experiments. As shown in TABLE~\ref{table: missing data}, GFTD demonstrates competitive performance under conditions of incomplete data. Even when $75\%$ of the historical observations are missing, the average JADE/JFDE performance only declines by $8.9\%$ and $7.4\%$, respectively, still surpassing S-GAN with complete data input by a significant margin. In contrast, Joint AgentFormer fails to produce plausible predictions under high missing data rates.

\section{Conclusion \& Discussion}
In this work, we present the Guided Full Trajectory Diffuser, a novel framework for representing the joint distribution of trajectories leveraging diffusion models, converting trajectory prediction and controllable generation into a unified inverse problem. We formulate the prediction task as spatial-temporal inpainting, a general noisy inverse problem that can be solved through diffusion posterior sampling. Under this framework, the generated trajectories are not rigidly constrained by their historical observations; instead, we gradually enforce such conditions by adjustable posterior guidance. Such a unique design enables flexibility, resulting in not only competitive performance in joint trajectory prediction tasks but also generalizable to scenarios with noise perturbation or incomplete historical data. Moreover, our framework is compatible with plug-and-play modules and various guidance methods, thus expandable to task-oriented enhancements. However, we point out that our current implementation only considers raw trajectories as guidance reference, how to more efficiently exploit the interaction to further improve guidance quality is a topic we will keep working on.

\bibliographystyle{unsrt}
\bibliography{main}
\end{document}